\documentclass[]{spie}  

\usepackage{amsmath,amsfonts,amssymb}
\usepackage{graphicx}

\usepackage{url}            
\usepackage{booktabs}       
\usepackage{amsfonts}       
\usepackage{nicefrac}       
\usepackage{microtype}      
\usepackage{subcaption}

\usepackage{times}
\usepackage{epsfig}
\usepackage{graphicx}
\usepackage{amsmath}
\usepackage{amssymb}
\usepackage{blindtext}
\usepackage{fixltx2e}
\usepackage{algorithmic}
\usepackage{algorithm2e}
\usepackage{float} 
\usepackage{lscape}
\usepackage{multicol}
\usepackage{tabularx} 

\title{A Sparse Coding Multi-Scale Precise-Timing Machine Learning Algorithm for Neuromorphic Event-Based Sensors}

\author[a]{Germain Haessig}
\author[abc]{Ryad Benosman}
\affil[a]{Sorbonne Universite, INSERM, CNRS, Institut de la Vision, 17 rue Moreau, 75012 Paris, France.}
\affil[b]{University of Pittsburgh Medical Center, Biomedical Science Tower 3, Fifth Avenue, Pittsburgh, PA.}
\affil[c]{Carnegie Mellon University , Robotics Institute, 5000 Forbes Avenue Pittsburgh PA 15213-3890.}

\authorinfo{Further author information: (Send correspondence to Germain Haessig)\\Germain Haessig: E-mail: germain.haessig@inserm.fr, \\  Ryad Benosman: E-mail: ryad.benosman@upmc.fr}

\begin{document} 
\maketitle

\begin{abstract}
This paper introduces an unsupervised compact architecture that can extract features and classify the contents of dynamic scenes from the temporal output of a neuromorphic asynchronous event-based camera. Event-based cameras are clock-less sensors where each pixel asynchronously reports intensity changes encoded in time at the microsecond precision. While this technology is gaining more attention, there is still a lack of methodology and understanding of their temporal properties. This paper introduces an unsupervised time-oriented event-based machine learning algorithm building on the concept of hierarchy of temporal descriptors called time surfaces. In this work we show that the use of sparse coding allows for a very compact yet efficient time-based machine learning that lowers both the computational cost and memory need. We show that we can represent visual scene temporal dynamics with a finite set of elementary time surfaces while providing similar recognition rates as an uncompressed version by storing the most representative time surfaces using clustering techniques. Experiments will illustrate the main optimizations and trade-offs to consider when implementing the method for online continuous vs. offline learning. We report results on the same previously published 36 class character recognition task and a 4 class canonical dynamic card pip task, achieving $100\%$ accuracy on each.
\end{abstract}


\section{INTRODUCTION}
\label{sec:intro}  

Neuromorphic event-driven time-based vision sensors operate on a very different principle than conventional frame-based cameras. Instead of acquiring static images of a scene, these sensors asynchronously record pixel intensity changes with a high temporal precision (around $1\mu s$). The event format differs significantly from frames, and therefore conventional machine learning algorithms cannot be directly applied if one wants to fully use its potential and temporal properties in terms of power consumption, computational cost and low memory requirements. 
 Previous notable work on object recognition using event-driven time-based vision sensors include real-time event-driven visual pattern recognition that recognizes and tracks circles of different sizes using a hierarchical spiking network running on custom hardware \cite{CAVIAR2009}, a card pip recognition task on FPGAs, implementing different hierarchical spiking models inspired by Convolutional Neural Networks (CNNs) \cite{MappingPerez-Carrasco2013} and new methods such as HFirst \cite{orchard2015hfirst} and recently HOTS \cite{lagorce2015hots} , an unsupervised algorithm which fully considers the spatio-temporal aspects of event-based sensors.
In this paper we introduce a compact hierarchical event-driven multi-temporal framework to learn spatiotemporal patterns of dynamic scenes extending the concept of hierarchically increasing spatiotemporal-scales time-surfaces introduced in \cite{lagorce2015hots} .
A time-surface is a descriptor that provides a time context around an incoming event and describes the temporal activity in its surrounding.
This descriptor, applied in a multilayer architecture requires a high number of features to correctly characterize a dynamic visual input and therefore high memory and computational costs. Ideally there must be a one to one relation between interesting spatio-temporal time-surfaces detected in the scene and the ones stored in each layer of the architecture.
The aim of this work is to present a new formulation of the method in order to reduce the number of features using sparse coding to express any time-surface as a linear combination of elementary time-surfaces rather than selecting the most representative ones using clustering techniques (iterative K-means \cite{barbakh2008online} in the original paper).

\section{Event-based cameras}
\label{section::atis}
\begin{figure}[htbp]
\centering
\includegraphics[width=0.6\textwidth]{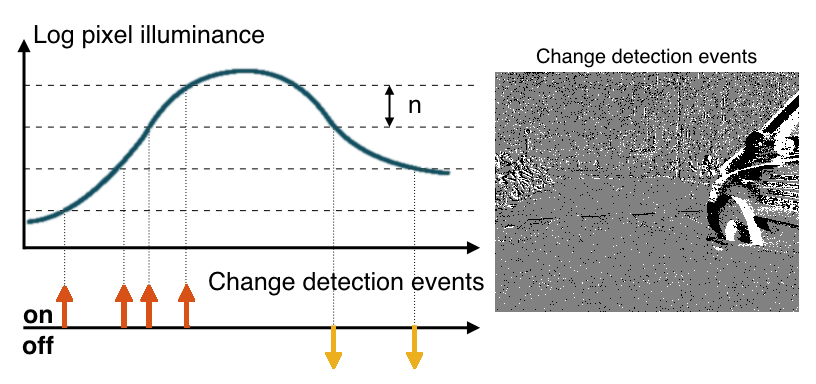}
\caption{a) The ATIS sensor. b) Functional diagram of an ATIS pixel. Two types of asynchronous events, encoding change and brightness information, are generated and transmitted individually by each pixel in the imaging array. Each time the luminance of the considered pixels rises a level (a relative change from previous measurement), a change detection event is generated, like at time $t_0$. This triggers a second circuitry which measures the absolute gray level value. This value is coded in the timing between two events occurring at $t_1$ and $t_2$. This timing encodes the time required to capture a given amount of light, and is then inversely proportional to the absolute gray level.
\label{fig:atisprinciple} }
\end{figure}

Biomimetic event-based cameras are a novel type of vision sensor that are event driven. Unlike their frame-based counterparts, they are not controlled by artificially created timing and control signals (frame clock) with no relation to the source of the visual information. Events are generated when significant changes of the relative luminance occur at the pixel level as shown in Figure.\ref{fig:atisprinciple}. The visual output is in the form of an Address Event Representation and encodes the visual information in the time dimension at the microsecond time precision. As soon as a change of luminance is detected, the process of communicating the event off chip is initiated. The process executes with low latency, of the order of a microsecond, ensuring that the time at which an event is read out from the camera inherently represents the time at which a contrast change was detected.
The camera used in our setup is an Asynchronous Time-based Image Sensor (ATIS) \cite{posch2011qvga} which has a $304 \times 240$ pixel resolution. This array of fully autonomous pixels combines both a relative luminance change detector circuit and a conditional exposure measurement block. Only the change detector circuit will be used for our experiments. When no change of luminance is detected, no events are generated and the static information is not recorded. This reduces the data load and allows for high speed online processing. 

\section{Methods}
In this section, we will introduce the notion of time-surface, describing the spatio-temporal surrounding of an event, before showing that these so defined time-surfaces can be clustered using a sparse coding algorithm, thus allowing a drastic reduction of the total number of features used for classification, which directly impacts the hardware implementability of the algorithm, as discussed in next section.
\subsection{Time-surface construction}
\label{section::time_surface}

\begin{figure*}[h]
\centering
 \includegraphics[width=0.9\textwidth]{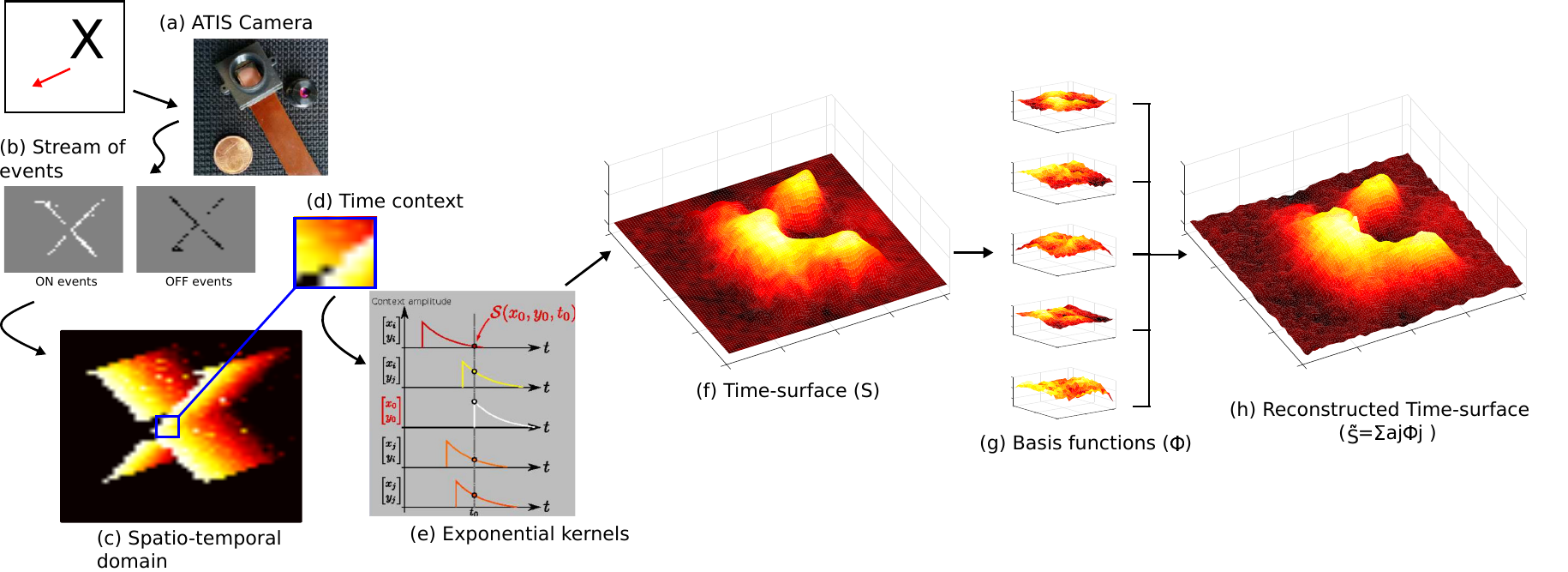}
 \caption{Time-surface generation and decomposition over a basis. The ATIS camera (a) generates a stream of events (b) from an moving X character. The time context (d) of one event is extracted from the spatio-temporal domain (c) and convolved with and exponential decay (e), providing a time-surface (f). This time-surface is projected over a set of basis functions (g), which gives the resulting (h) reconstruction. Inspired from \cite{lagorce2015hots} .}
\label{fig:TSgeneration}
\end{figure*}

We consider a stream of visual events (Figure \ref{fig:TSgeneration}.b), which we define as $ev_i=\{x_i,y_i,t_i, p_i\}$ where $ev_i$ is the i-th event, and consists of its spatial location ($x_i$, $y_i$), its timestamp $t_i$, and its polarity $p_i$, with $p_i$  $ \in $ \{-1, 1\}, for respectively a decrease or increase in luminance.\\
In \cite{lagorce2015hots} the notion of time-surface feature $S_i$ is introduced. It represents the previous activity around the spatial location of an incoming event $ev_i$. Thus, for an incoming event $ev_i$, we define a time context (Figure \ref{fig:TSgeneration}.d) as an array of the last activation time in the $(2R+1)^2$ neighborhood (of radius $R$), centered at ($x_i$, $y_i$).\\
The time-surface $S_i$ (Figure \ref{fig:TSgeneration}.f) around an incoming event $ev_i$ is then obtained by applying an exponential decay (Figure \ref{fig:TSgeneration}.e) to each element of the time context \cite{lagorce2015hots}) .\\
For an incoming event $ev_i$ at location ($x_i$,$y_i$), the value of its associated time-surface is:
\begin{equation*}
\label{eq:surf}
S_i(u,v) = e^{-(t_i-t_{u,v})/\tau}  
\end{equation*}
where $u  \in [\![ x_i-R, x_i+R]\!]$, $v \in  [\![ y_i-R, y_i+R  ]\!]$, $t_{u,v}$ the timestamp of the most recent event that occurred at the respective pixel, $t_i$ the timestamp of the current event and $\tau$ the time constant of the exponential decay. As all the events are processed one after the other, the time difference $t_i-t_{u,v}$ is necessary positive. Figure \ref{fig:TSgeneration}.b shows this full process of time-surface generation.

\subsection{Training phase: Finding the patch of projection basis}
\label{section::projection}
Following the sparse coding algorithm introduced in \cite{olshausen1996emergence} , assuming that any time-surface $S(x, y)$ can be approximated as a linear combination of $N$ functions $\phi_j(x, y) $ for $j \in [\![ 1, N]\!]$ giving its estimation $\tilde{S}$ (for clarity, we omit the $i$ subscript for each event) :
\begin{equation*}
\label{eq:emc} 
\tilde{S}(x, y)= \sum_{j=1}^{N}  a_j \phi_j(x, y)  
\end{equation*}
where the $a_j$ are real linear coefficients ($a_j\in \mathbb{R} $ for $j \in [\![ 1, N]\!]$).
This relation is equivalent to the projection of a time-surface $S$ onto a subspace defined by the elementary time-surfaces $\phi_j$.
The feature estimation can be classically formulated as an optimization problem.  The solution is given by minimizing the following residual error function $E$, set as the difference between the original surface and its reconstruction using the elementary time-surfaces linear summation (left member) and a measure of the sparseness of the coefficients (right member):
\begin{equation}\label{eq:basis update}
E = \underbrace{ \sum_{x,y} \left[ S(x, y)- \sum_{j=1}^{N}  a_j \phi_j(x, y)  \right]^2 }_{\textit{ reconstruction error}} + \underbrace{ \lambda \sum_{j=1}^N|\frac{a_j}{\sigma}|}_{\textit{ sparseness of the ($a_j$)}}
\end{equation}
Where $\lambda$ is a positive constant that determines the influence of the second term relative to the first, and $\sigma$ a scaling constant. \\

The minimization is performed using a training dataset of events from the event-based camera \cite{olshausen1996emergence} .
Elementary time-surfaces are initialized with random uniformly distributed values between $0$ and $1$, while the coefficients values are initialized with the product between the initial features and the time-surfaces computed from the training dataset. We use the conjugate gradient descent method \cite{press1988numerical} to minimize the error $E$ (i.e. maximize the similarity between $S$ and $\tilde{S}$, by updating the coefficients $a_j$, while maximizing the sparseness of the coefficients). \\
Elementary time-surfaces are then updated by adding the residual error weighted by the coefficients obtained from the previous iteration of the minimization process: 
\begin{equation*}
\phi_j \longleftarrow  E.a_j.\eta + \phi_j
\end{equation*}
where $\eta$ is the learning rate. All time-surfaces are computed offline from the learning dataset. Then, they are fed to the minimization algorithm until convergence.
The number of features is computed iteratively and set as the one providing the lowest reconstruction error on the training time-surfaces set.

\subsection{Building a hierarchical model}
\label{section::hierarchy}

In the initial architecture described in \cite{lagorce2015hots} , each layer has a set of elementary time-surface prototypes matching time surfaces from the observed scenes and learned during the training phase. When an input time-surface matches a prototype, an event is produced and transmitted to the next layer. A single input event will produce at most one output event. In the present model, a layer now contains a finite set of $N$ elementary time-surfaces while the output of a layer represents the linear response of all the $N$ elementary time-surfaces, represented by the projection coefficients of the input surface onto this basis. 

\begin{landscape}
\begin{figure}[h]
\centering
 \includegraphics[width=1.2\textwidth]{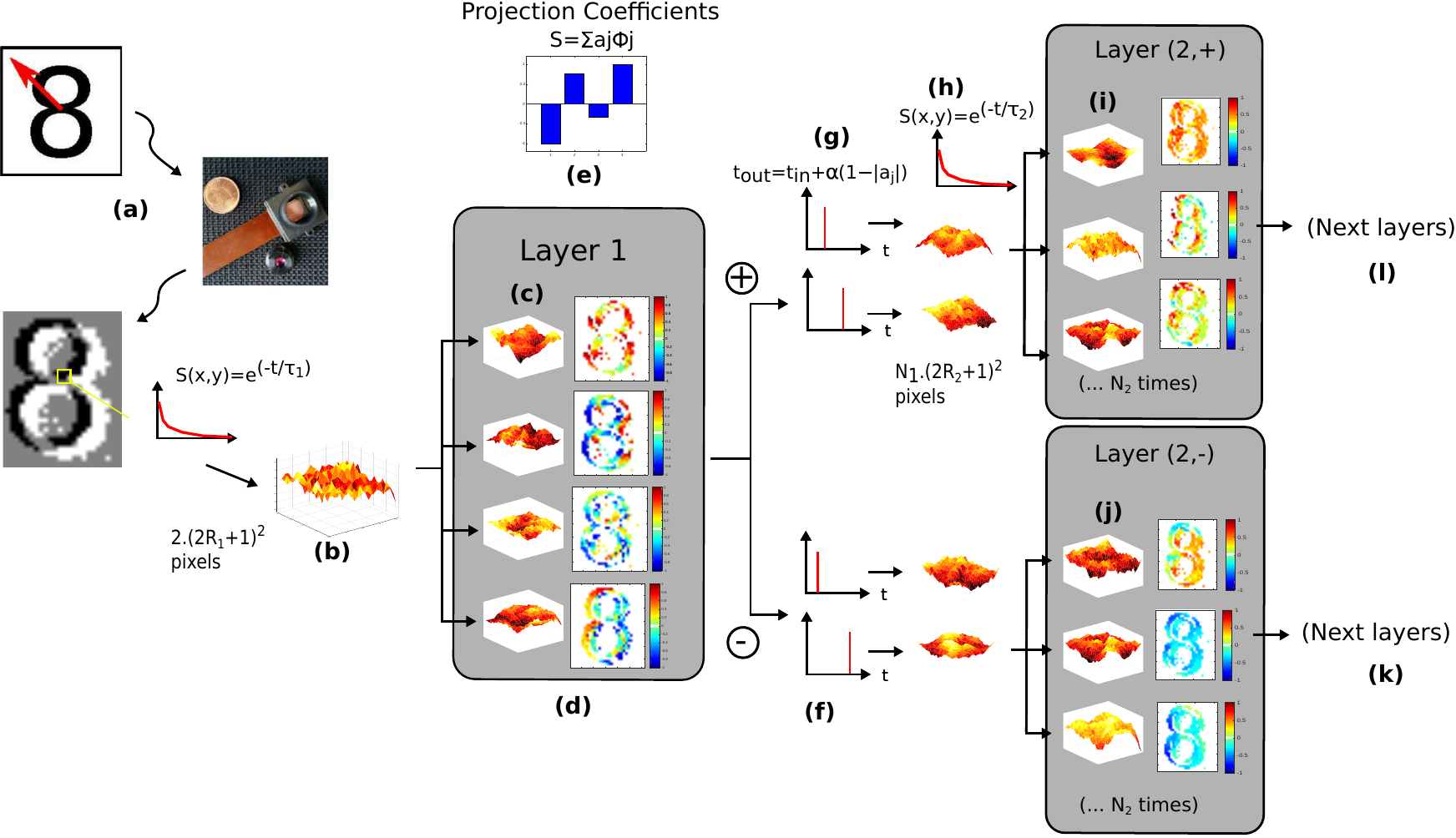}
 \caption{
Detailed view of a two layer architecture : \textbf{(a)} A moving object is presented to the ATIS camera which outputs a stream of events. \textbf{(b)} Each event $ev_{in}$ is acquired and a time surface is computed using a $\tau_1$ time constant. As the event stream here contains two polarities (ON and OFF), the time-surface contains the two polarities and is of size $2 \times (2R_1+1)^2$. \textbf{(c)} Time-surfaces are presented to layer 1 and an elementary set of $N_1$ prototypes \textbf{\textit{(cf. eq \ref{eq::param_evolution})}} is extracted. \textbf{(d)} After learning this elementary set, recorded time-surfaces are presented again to Layer 1, and projected onto all prototypes. \textbf{(e)} The projection coefficients are obtained as the least square solutions that minimizes the error function $E$ \textbf{\textit{(cf. eq \ref{eq:basis update})}}. \textbf{(f,g)} All the projection coefficients are split between positive and negative values. The projection coefficients are convolved with a linear decay, and the results are used to determine the spike times $t_{out}$ of the next emitted events $ev_{out}$ \textbf{\textit{(cf. eq \ref{eq::ev_new})}}. \textbf{(h)} Timestamps are sorted in ascending order and convolved with an exponential decay of integration time $\tau_2$ \textbf{\textit{(cf. eq \ref{eq::param_evolution})}} to obtain new time-surfaces, of size $N_1 \times (2R_2+1)^2$. \textbf{(i)} Time-surfaces are presented to layer $(2,+)$\textbf{(i)} and $(2,-)$\textbf{(j)} and a projection basis of $N_2$ \textbf{\textit{(cf. eq \ref{eq::param_evolution})}} elementary time-surfaces is estimated. \textbf{(k, l)} The same process can be repeated for the next layers, building in this way the proposed hierarchical model. Inspired from \cite{lagorce2015hots} .
 }
 \label{fig:hierarchy}
\end{figure}

\end{landscape}

Figure \ref{fig:hierarchy} shows the response obtained at the output of a layer, and how it is sent to the next layers in order to build the proposed hierarchical model. Figure \ref{fig:hierarchy}.a shows an object moving in front of an ATIS camera. Each event is captured from the stream sent by the event-based camera. In Figure \ref{fig:hierarchy}.b, time-surfaces for each event are built by a convolution with an exponential kernel of time constant $\tau_1$, for a neighborhood of ($2R_1+1$) side length. As described in section II, time-surfaces are sent to Layer 1, to find a projection basis of $N_1$ elementary time-surfaces shown in Figure \ref{fig:hierarchy}.c. Once the elementary time-surfaces projection basis has been extracted, the learning process of Layer 1 is finished.
Now each incoming time-surface is directly projected onto all elementary time-surfaces basis.  
The projection coefficients (Figure \ref{fig:hierarchy}.e) are defined as the least square solutions that minimize the error function $E$ (Equation \ref{eq:basis update}). The $a_j$ coefficients are constrained  between -1 and 1. The projection coefficients are split into two groups (Figure \ref{fig:hierarchy}.f-g), depending on whether their value is positive or negative, as detailed in \cite{lagorce2015stick} .
For each feature $j$ of the basis, a linear decay is applied to the projection coefficients as shown in Figure \ref{fig:hierarchy}.g, and a new event $ev_{out}$ is generated :
\begin{equation*}
ev_{out}=\{x_{out},y_{out},t_{out}, p_{out}\}
\end{equation*}
where :
\begin{align}
\begin{cases}
 (x_{out},y_{out}) &= (x_{in},y_{in}) \\
t_{out} &= t_{in}+\alpha \left( 1 - |a_j| \right) \\
p_{out} &= j \hspace{0.5cm} | \hspace{0.5cm} j \in [\![ 1, N_1]\!]\\
\end{cases}
\label{eq::ev_new}
\end{align}
were $\alpha$ is a timing scale factor describing the time span of the newly generated event-stream. The higher the coefficient, the higher the similarity between the basis and the surface input. We then encode the similarity level into the time domain, as a delay of the output event depending on the similarity between two surfaces. \\
As shown in Figure \ref{fig:hierarchy}.h, new events are convolved with an exponential decay of time constant $\tau_2$, for a neighborhood of ($2R_2+1$) side length, in order to generate new time-surfaces. In Figure \ref{fig:hierarchy}.i, output delayed spikes, for positive coefficients, according to the results of convolutions with each elementary time-surfaces basis are sent to Layer $(2,+)$ for training. The goal is now to determine the $N_2$ elementary time-surfaces of Layer 2. As shown in Figure \ref{fig:hierarchy}.j, steps (d) to (h) are repeated. Then, the same steps (g) and (h) are applied to Layer $(2,-)$ for negative coefficients.

Since the nature of the input and the output of our model is the same, events produced at the output of Layers $(2,+)$ and $(2,-)$ can be sent to the next layer, and the process from step (b) to (g) is repeated (Figure \ref{fig:hierarchy}.l-k). Layers are trained consecutively one after the other.

As in \cite{lagorce2015hots} , the main idea of the architecture is to gradually increase the complexity of spatial and temporal information. Information is integrated over larger and larger time scales. 
Each layer $L_i$ increases its number of elementary time-surfaces $N_i$, the neighborhood radius $R_i$, and the integration time constant $\tau_i$:
\begin{align}
\begin{cases}
\tau_{i+1} &= K_{\tau} \tau_i \\
R_{i+1} &= K_{R}R_i \\
N_{i+1} &= K_{N}N_i \\
\end{cases}
\label{eq::param_evolution}
\end{align}
The network is only defined by a set of six parameters : the initial conditions $\left(\tau_0, R_0, N_0\right)$ and the evolution parameters $\left( K_{\tau}, K_{R} ,K_{N} \right)$.
Comparisons with \cite{lagorce2015hots} are provided in the experiments section.

\subsection{Classification}
\label{section::classification}
\begin{figure}[h]
    \centering
    \begin{subfigure}[b]{0.4\textwidth}
 \includegraphics[width=\textwidth]{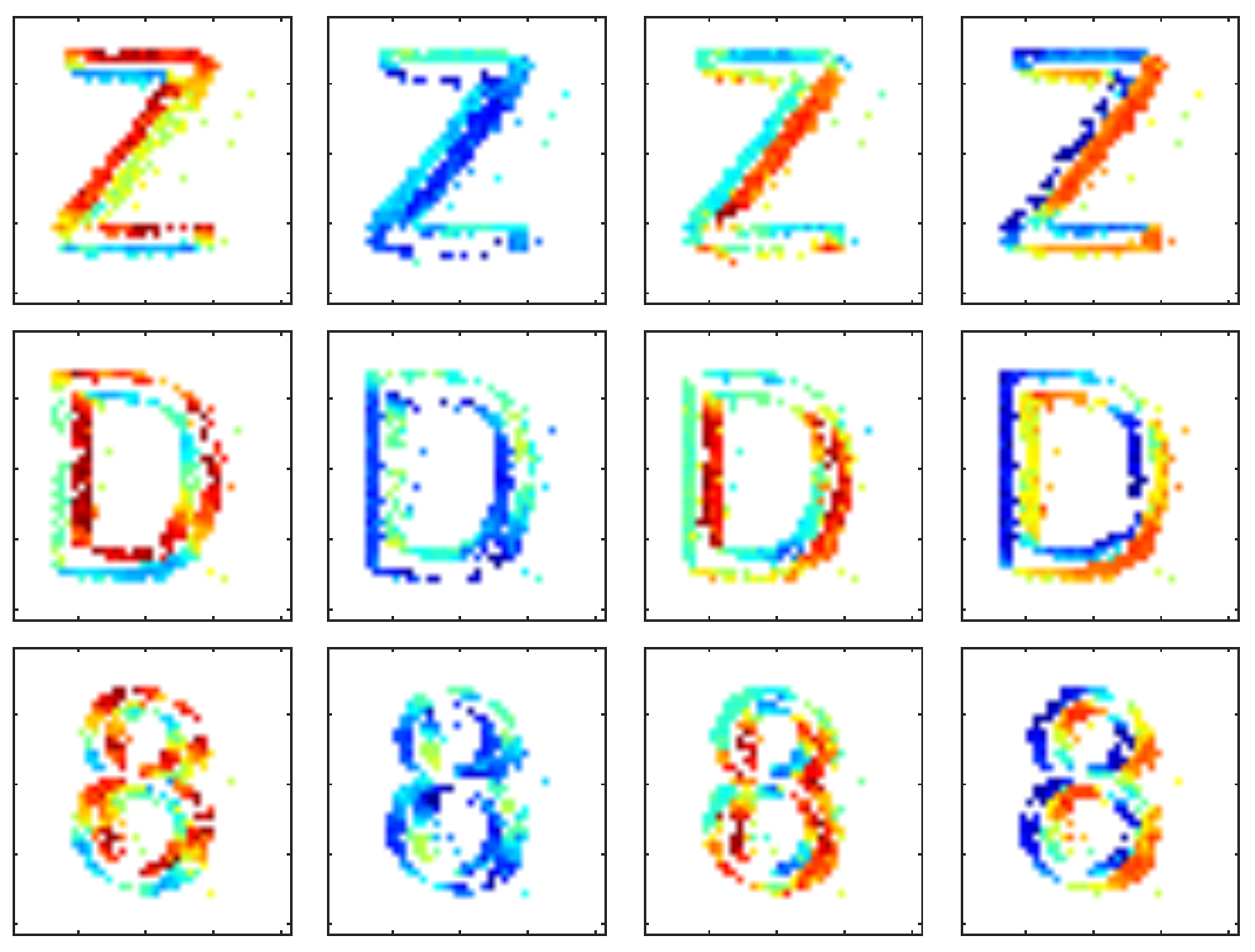}
 \caption{}
 \includegraphics[width=\textwidth]{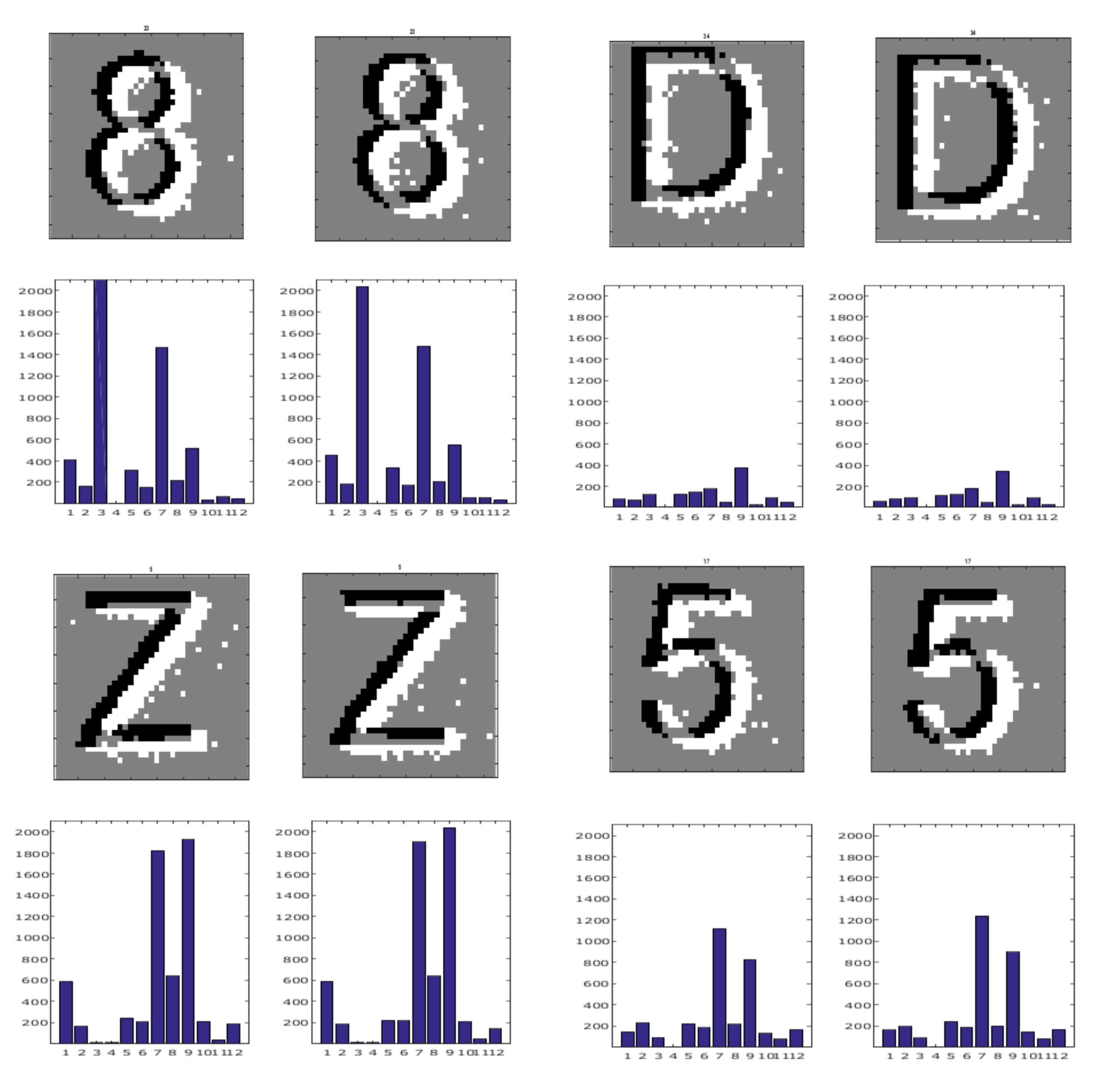}
  \caption{}
  \label{fig:out_char}
    \end{subfigure}
    ~ 
    \begin{subfigure}[b]{0.56\textwidth}
          \includegraphics[width=\textwidth]{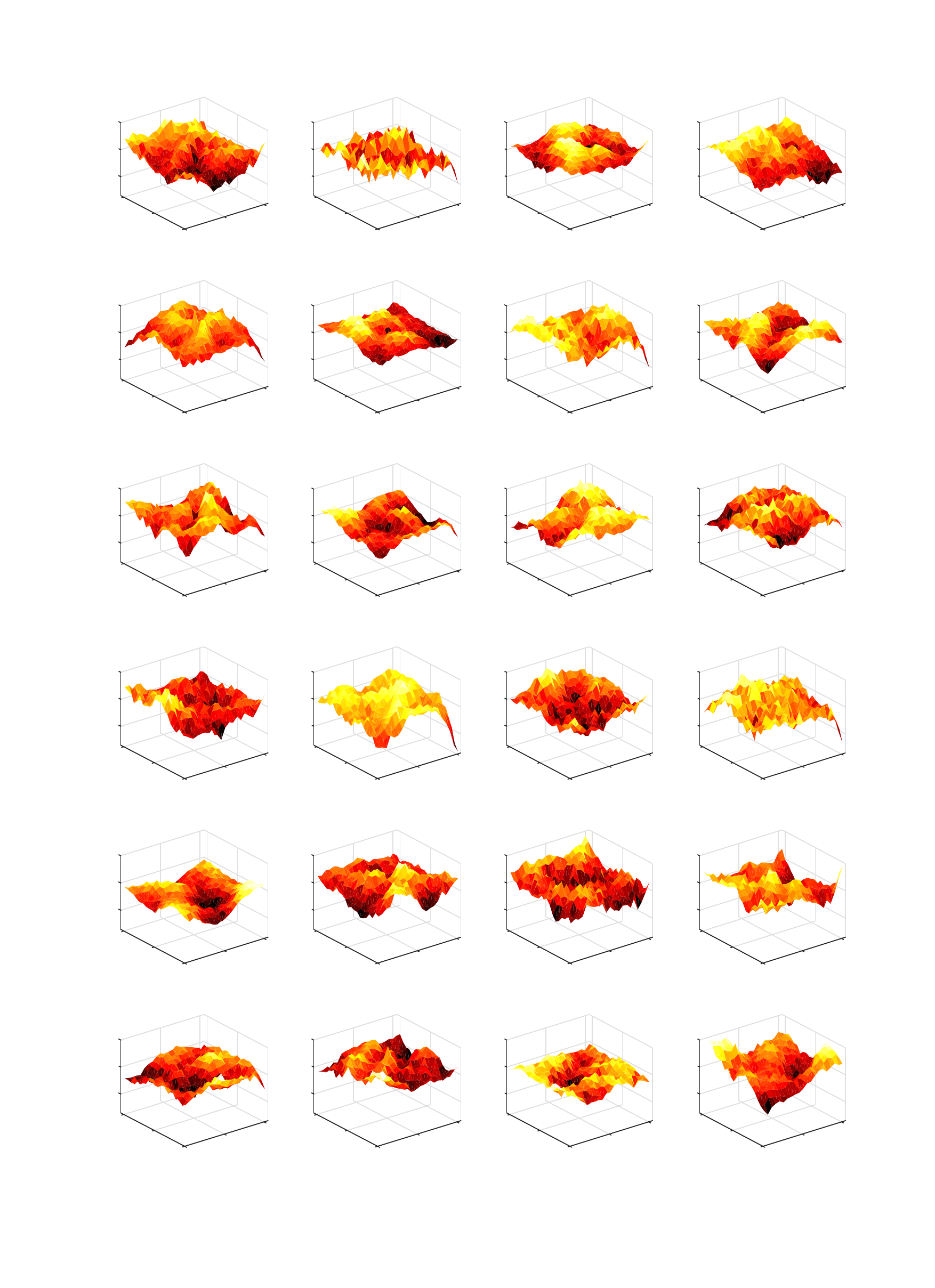}
  \caption{}
  \label{fig:64bases}
    \end{subfigure}
    \caption{Results for the character dataset. a) Output of the 3 layers network for three characters after learning. This output is color coded according to the responding basis (1 out of 24). We can see that some basis are specialized (red for horizontal edges, dark blue for curves, etc). b) Signature for two different representations of the same digit. The signatures are very similar between the same class and differ from class to class, allowing a good classification. c) 24 bases of the third layer of the model after learning on the letters and digits dataset. Because they are combination of responses of the previous layer, they are difficult to interpret as they are, but one can notice that they differ one from the other.}
\end{figure}

The output of the last layer of the hierarchical structure is fed to a classifier for pattern classification. We use a simple classifier to show that the model provides sufficient discrimination without the need for complex classification methods. 
However, in case of larger databases, where a variety of descriptors are generated for the same class, the use of more advanced classification algorithms might become necessary.
We compute an histogram from the output of the last layer of the hierarchical model that contains the total number of responses of each feature to the input pattern, independently from its spatial position. This is the signature of the observed object, and will be used for further comparisons. This method is the same as the one explained in \cite{lagorce2015hots} . Figure \ref{fig:out_char} shows an example of such an histogram.

Each learning example is presented to the model, in order to learn its signature.
In a second step, during the testing phase, the signature of incoming patterns are computed and compared to the signatures of each learned examples. The closest one is then identified and finally, the recognized object is obtained by a majority vote from the results of all the last sub-layers.

We will use two types of distances between signatures : the Euclidean distance and the Bhattacharyya distance \cite{bhattacharyya1946measure} . The histogram signature for classification, which is a weak classifier, is here chosen in order to prove the robustness of the proposed descriptor. Stronger classifiers exist (SVM, adaboost methods, ...), but their use in this context is outside the focus of this paper.

\section{Experiments}
\label{section::experimentation}

\begin{figure}[h]
\centering
  \includegraphics[width=0.7\textwidth]{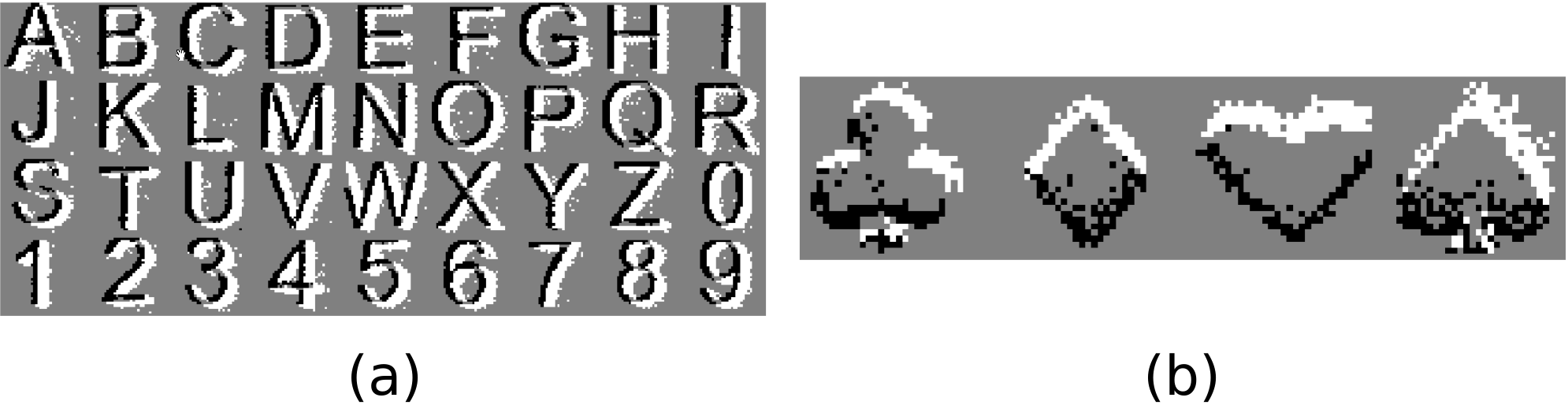}
  \caption{Representation of the datasets used for the experiments: (a) Letters and digit dataset: consisting of the 26 characters of the roman alphabet and the digits from  0 to 9. (b) Flipped cards dataset: consisting of the four suits of a deck of cards (club, diamond, heart and spade). Inspired from \cite{lagorce2015hots} .}
\label{fig:datasets}
\end{figure}

\subsection{Letters and digits dataset}
The model has been used with the "Letters and Digit" dataset, provided by Orchard et. al \cite{orchard2015hfirst} (Figure \ref{fig:datasets}-a). It consists of a set of moving characters.  
The dataset used for the learning phase contains the representation of 26 characters from the roman alphabet and ten digits (A-Z, 0-9). The dataset used for the testing phase contains 12 representations of each character, in  [A-Z] and [0-9].
The goal is to identify the character or the digit.
For this experiment, we tested our method on a three layer architecture, we set empirically the number of elementary time-surfaces for each layer to be respectively: $N_1=6$, $N_2=9$, $N_3=12$.

The neighborhood radius is empirically set to $R=2$ throughout the whole architecture, while the integration times for the exponential decays are set to : $\tau_1=10$ms, $\tau_2=15$ms and $\tau_3=20$ms.
We were able to obtain 100\% of accuracy in classification, both with the Euclidean and the Bhattacharyya distance.

\subsection{Flipped card deck}

The second dataset we used is the "Flipped card deck" dataset, provided by Linares-Barranco et. al \cite{perez2013mapping} (Figure \ref{fig:datasets}-b). It consists of a deck of cards whose corner is flipped in front an event-based camera so that only the suit symbol of the card is visible. The goal is to identify the cards' suit. The original dataset contains 40 samples, 10 from each suit. The learning and testing datasets were generated by randomly taking respectively 7 and 3 examples of each suit from the original dataset. For this experiment, we tested our method on a three layer architecture, with the same parameters used for the letters and digits experiment. We were able to obtain a 100\% accuracy. We then decreased the number of features for each layer, in order to see the impact of such a modifications. For $(N_1,N_2,N_3) = (3,6,9)$, the recognition rate drops to 83\%, while decreasing the number of generated spikes by a factor of $2.5$ (see Table \ref{tab:all}).

\begin{table}[h]
\caption{Results and comparison with original HOTS model}
\centering
\begin{tabular}{|c|c|c|c|c|c|}
\hline
 Algorithm   & \multicolumn{3}{c|}{This paper}  & \multicolumn{2}{c|}{HOTS \cite{lagorce2015hots}}    \\ \hline
  Dataset      & \multicolumn{2}{c|}{Cards} & Digits & Cards & Digits \\ \hline \hline
Number of Centers & 3-6-9 & 6-9-12 & 6-9-12 & 8-16-32 & 8-16-32 \\ \hline
Recognition Rate & 83\%& 100\% & 100\% & 100\% & 100\% \\ \hline 

Number of spikes & 7 599 450 & 18 500 730 & 74 440 535 & 52 410 & 513 383   \\ \hline 
\end{tabular}
\label{tab:all}
\end{table}  

\section{Conclusion}
This work exposes a new approach for learning spatiotemporal features from an event based camera. It allows a compact representation of information by reducing the number of prototypes in a hierarchical  learning structure using sparse coding basis decomposition. 
This reduction induces an increase in the number of events generated by the system. Unlike the original model, we have shown that it is possible to reduce the number of prototypes without strongly impacting the precision.
Our model increases the activity of each cell. This, however, is not a limitation since the available neuromorphic hardware can handle a large number of spikes (signal constraints) despite being limited to a small number of neurons (hardware constraints). Limitations are then shifted to maximal firing rate and bandwidth, not to mapping constraints due to a lack of available neurons. It is important to emphasize that we assumed a floating point precision, no consideration was given to precision limitations.
Future work will focus on characterizing a loss of precision, both in number representation and in basis projection errors. Moreover, the addition of a refractory period to each prototype, in order to limit the firing rate of a single cell, will be studied. All of this could lead to a full hardware implementation.

\acknowledgments 
 
The authors want to thanks Camille Simon Chane for a careful verification of the manuscript. 

\bibliography{main} 
\bibliographystyle{spiebib} 

\end{document}